# PERFORMANCE ANALYSIS OF UNSYMMETRICAL TRIMMED MEDIAN AS DETECTOR ON IMAGE NOISES & ITS FPGA IMPLEMENTATION

K.Vasanth[1] and S.Karthik[2]

[1]Department of Electrical & Electronic Engineering, Sathyabama University, Chennai, Tamilnadu, India
vasanthecek@gmail.com
[2]Project Associate, Cognizant Technology Solutions, Chennai, Tamilnadu, India
skarthick76@gmail.com

## ABSTRACT

*This Paper Analyze the performance of Unsymmetrical trimmed median, which is used as detector for the detection of impulse noise, Gaussian noise and mixed noise is proposed. The proposed algorithm uses a fixed 3x3 window for the increasing noise densities. The pixels in the current window are arranged in sorting order using a improved snake like sorting algorithm with reduced comparator. The processed pixel is checked for the occurrence of outliers, if the absolute difference between processed pixels is greater than fixed threshold. Under high noise densities the processed pixel is also noisy hence the median is checked using the above procedure. if found true then the pixel is considered as noisy hence the corrupted pixel is replaced by the median of the current processing window. If median is also noisy then replace the corrupted pixel with unsymmetrical trimmed median else if the pixel is termed uncorrupted and left unaltered. The proposed algorithm (PA) is tested on varying detail images for various noises. The proposed algorithm effectively removes the high density fixed value impulse noise, low density random valued impulse noise, low density Gaussian noise and lower proportion of mixed noise. The proposed algorithm is targeted on Xc3e5000-5fg900 FPGA using Xilinx 7.1 compiler version which requires less number of slices, optimum speed and low power when compared to the other median finding architectures.*

## 1. INTRODUCTION

Images are often corrupted by noises due to poor image sensors or error in transmission medium. The different types of noises that occur in images are additive random noise such as Gaussian white noise and salt-and-pepper impulse noise, signal-dependent noise such as speckle [1]. In order to restore the corrupted images, a suitable filter should be used. A good noise removal filter would exactly restore the image by removing the noise distributions only. So to obtain the above result, a suitable filtering algorithm must be stated to remove a noise distribution. In Practise, the noise removal filter is designed to restore images well but there will be always some degree of variation in the restored pixel values from the original image. If there is much deviation the chosen algorithm is not suitable for the restoration In the above stated condition the restored image might not be visually unacceptable if subjected to human inspection [2]. The poor photo electronic detectors which results in thermal noise which is modelled as additive zero mean Gaussian noise corrupts the images [3]. Impulse noise is caused by transmission in a noisy channel. Basically there are two common types of impulse noise are the salt-and-pepper noise and the random-valued noise. For images corrupted by salt-and pepper noise, the noisy pixels can take only the peak and the valley values while in the case of random-valued noise; they can take any random value in the dynamic range [3]. A conventional method to remove noise from image data is to use a spatial filter. Spatial filters broadly classified into non-linear and linear filters. Many non-linear filters fall into the category of order statistic neighbourhood operators. This means that the local neighbours are ordered in ascending order and this list is processed to give an estimate of the underlying image brightness. The simplest order statistic operator is the median [3], where the central value in the ordered list is used for the new value of the brightness. The median is good at reducing impulse noise However, A mean or average filter is the optimal linear filter for Gaussian noise removal which tends to blur sharp edges, destroy lines and other fine image details. Median filter often blur the image for larger window size and insufficient noise suppression for small window sizes [4]. Adaptive Median Filter (AMF) blurs the image at high noise densities but fairs well at low and medium noise densities [5].

In Threshold decomposition filter (TDF) the pixels are decomposed based on various threshold levels and subjected to Boolean operation. This eliminated the need for complex sorting technique. This decomposition algorithm requires large threshold levels for operation and fails at higher noise densities. The above mentioned Median and its variant filters operate uniformly over the entire image results in the modification of uncorrupted pixel. Ideally the filtering should be applied only to corrupted pixels while leaving uncorrupted pixels intact. Therefore, a noise-detection process should discriminate between uncorrupted pixel and the corrupted pixel prior to applying nonlinear filtering is highly desirable. To elude the drawback of the above filters switched median filters were introduced. These filters work on the basis of impulse detection and correction. One of the popular switched median filter is Progressive Switched Median filter (PSMF). In this filter the decision is based on fixed threshold value and hence a procuring a strong decision is difficult. Hence at increasing noise densities the switched filters do not consider any of the local detail of the image and hence edges are not preserved properly [6]-[7].

The DPF filter removes noise at medium noise densities but fails to eliminate salt and pepper noise at high noise densities [8]. Decision based filter [9] identifies the processed pixel as noisy, if the pixel value is either 0 or 255; else it is considered as not noisy. Under High noisy environment the DBA filter replaces the noisy pixel with neighbourhood pixel. Due to repeated replacement of neighbourhood pixel results in streaks in restored image. To avoid streaks in images an improved DBA (DBUTMF) [10] is proposed with replacement of median of unsymmetrical trimmed output, but under high noise densities all the pixel inside the current would take all 0's or all 255's or combination of both 0 and 255. Replacement of trimmed median did not fair well for above case. Hence Modified decision based un-symmetric trimmed median filter (MDBUTMF) [11] is proposed. The above cause is eliminated by replacing the mean of the current window. When the noise densities scale greater than 80% the Smudging of edges occurs. All the Estimation based Threshold algorithms and conventional algorithms fairs well for low and medium density impulse noise but fails at high noise densities also these algorithms do not preserve edges. Hence a suitable algorithm that detects, eliminates impulse noise and preserves edges for high noise densities is proposed. This paper is organized as follows. Section II describes noise model. Section III gives a overview of related work on Image De-noising using proposed algorithm and its hardware implementation. Section IV deals with Exhaustive Experimental Results and Discussions and finally Concluding Remarks are given in Section V.

## 2. NOISE MODEL

Let the true image *x* belong to a proper function space S(Ω) on Ω = [0; 1]$^2$, and the observed digital image *y* be a vector in R*mxm* indexed by A ={1,2,..m} X {1,2,.m}. The image degradation can be modeled as y = N(Hx), where H : S(Ω)→ R*mxm* is a linear operator representing blurring, and N : R*mxm* → R*mxm* models the noise. Usually, y = $Hx + \sigma n$ where σn Є R*mxm* is an additive zero-mean Gaussian noise with standard deviation σ>= 0 . Outliers are modeled as impulse noise. Then a realist model for our data is

$$y' = H.x.K + \sigma g \quad (1)$$

$$y = N(y') \quad (2)$$

Where *N* represents the impulse noise and K refers to speckle noise as given in equation 1 & 2. The noise model for salt & pepper noise is given below. If [0; 255] denote the dynamic range of *y'*, i.e., 0 <= y'ij <= 255 for all (*i,j*), then they are denoted by Salt-and-pepper noise: the gray level of *y* at pixel location (*i j*) is illustrated in the equation 3.

$$y_{ij} = \begin{cases} 0 & \text{with probability p;} \\ y'_{ij} & \text{with probability 1 - p - q;} \\ 255 & \text{with probability q;} \end{cases} \quad (3)$$

Where *s = p + q* denotes the salt-and-pepper noise level [12].

## 3. PROPOSED ALGORITHM

### 3.1. Snake like improved shear sorting

Over the years sorting algorithm is a basic operation behind all the median filters. All the existing sorting algorithms require more comparators. In this paper a new snake like improved shear sorting algorithm is proposed for ordering the entire array of processed pixels as shown in figure 1.

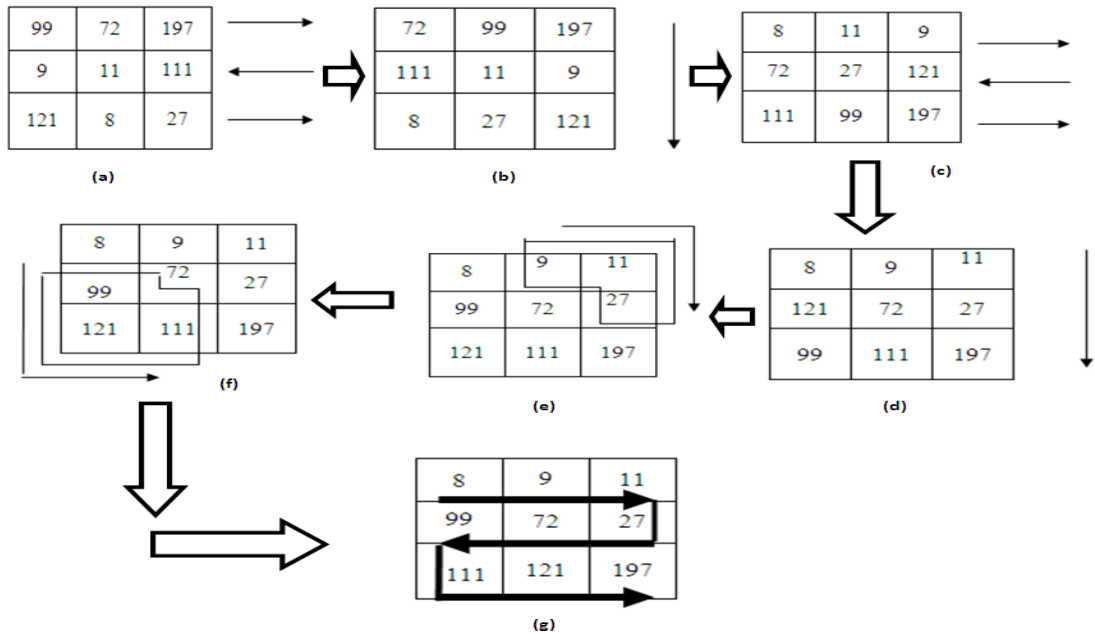

Figure 1 illustration of the proposed sorting methodology

Let D be an m x n matrix which is mapped with linear integer sequence W. Sorting the sequence W is then equivalent to sorting the elements of D in some Pre determined indexing scheme. The proposed Snake like modified algorithm consists of two basic operations row sorting, column sorting and semi diagonal sorting. The algorithm of the proposed snake like improved shear sorting algorithm is as follows.

Step1: The considered 2D processing window as shown in figure 1.a

Step2: Sort the $1^{th}$ and $3^{rd}$ rows of the 2D array in ascending order and $2^{nd}$ row in descending order independently. The sorted sequence is fed to step3 as shown in figure1.b.

Step3: Sort the three columns of the 2D array in ascending order. The sorted sequence is fed to step4 as shown in figure 1.c.

Step4: Repeat step 2 and 3 once again as shown in figure1.d and e.

Step5: Now Sort the upper semi diagonal of the semi sorted 2D array in ascending order as shown in figure1.e.

Step6: Sort the Lower semi diagonal sorted array in ascending order as shown in figure1.f. Resulting array is sorted in a snake like order. The procedure is repeated for the other windows of the image [13].

### 3.2. Proposed Algorithm

The brief illustration of the proposed algorithm is as follows.

Step 1: Choose 2-D window of size 3x3. The processed pixel in current window is assumed as $p_{xy}$.

Step 2: sort the 2D window data in ascending order using snake like modified shear sorting which is given by S. now Convert sorted 2D array into 1D array. $S_{med}$ is the median of the sorted array

Step 3: **Unsymmetrical trimmed median filter**

Initialize two counters, forward counter (F) and reverse counter (L) with 1 and 9 respectively. When a 0 or 255 are encountered inside the Sorted array (S), F is incremented by 1 or L is decremented by 1 respectively. The resulting array will be holding non noisy pixels of the current window. The median of this array is termed as UTMED (unsymmetrical trimmed median) [10].

Step 4: **Salt and pepper noise Detection**

Case (1): If the absolute difference between the processed pixel and unsymmetrical trimmed median filter (UTMED) is greater than the fixed threshold (T) then pixel is considered as noisy. As illustrated in equation 3

$$\text{If } |P(x,y)-UTMED| > T \quad (3)$$

Case (2): If the case 1 is true find the absolute difference between the median of and unsymmetrical trimmed median filter (UTMED). Check the difference is greater than the fixed threshold (T1) then median is considered as noisy as illustrated in equation 4. Case 2 is done for high noise densities where the computed median is also noisy.

$$\text{If } |S_{med}-UTMED| > T1 \quad (4)$$

Step 4: **Salt and pepper noise Correction logic**

If the case1 $|P(x, y)-UTMED| > T$ is true then check for the second case2 $|S_{med}-UTMED| > T1$. if both the condition are true then processed pixel and computed median is noisy. Hence replace the corrupted pixel with median of Unsymmetrical trimmed median. If condition 1 is true and condition 2 is false then corrupted pixel is replaced with the median of the sorted array. If both case 1 and case 2 fails then the pixel is termed as non noisy. The pixel is left unaltered [13].

### 3.3 Methodology of proposed work

The bigger matrix refers to image and values enclosed inside a rectangle is considered to be the current processing window. The element encircled refers to processed pixel. The above discussed methodology is illustrated as below.

$$\begin{pmatrix} 0 & 0 & 255 & 0 & 255 \\ 94 & 177 & 205 & 155 & 255 \\ 0 & 0 & 255 & 25 & 123 \\ 0 & 0 & 187 & 124 & 255 \\ 0 & 255 & 255 & 255 & 255 \end{pmatrix} \quad \begin{pmatrix} 0 & 0 & 255 & 0 & 255 \\ 94 & 177 & 205 & 155 & 255 \\ 0 & 0 & 155 & 25 & 123 \\ 0 & 0 & 187 & 124 & 255 \\ 0 & 255 & 255 & 255 & 255 \end{pmatrix}$$

Corrupted image segment　　　　Restored image segment

**Case (a):** Initialize forward counter F=1 and reverse counter L=9. Convert the 2D array into 1D array and sort the converted array. F and L counter moves in forward and reverse directions respectively. When a 0 is detected F is incremented by 1 and when a 255 is detected L is decremented by 1.

　Unsorted array:　177　0　0　205　255　187　155　25　124
　Sorted array $S_{xy}$　　0　0　25　124　155　177　187　205　255

Here the median $S_{med}$ value is 155. The case (1) is illustrated as follows. Now check for the presence of 0 or 255 in the sorted array. Every time a 0 is detected F is incremented by 1 and if 255 is detected L is decremented by1. In the above example there is two 0 and one 255. Hence F is incremented by two times and L is decremented by one time. Now finally F is holding 3 and L is holding 8. Now the variable DET is assigned with the median of the rank ordered unsymmetrical trimmed output i.e. corrupted pixel is replaced by median (25,124,155,177,187,205) = 166. i.e, DET=166. Now perform first step detection │255-166│ > 40. This condition is true. The Second condition is checked │155-166│ > 20 and the second condition is false. Hence the pixel is considered as noisy and median is considered as non noisy. The corrupted pixel is replaced by median of sorted array ie., output =155.

$$\begin{pmatrix} 0 & 0 & 255 & 0 & 255 \\ 94 & 177 & 0 & 0 & 125 \\ 0 & 0 & 185 & 0 & 255 \\ 0 & 0 & 255 & 255 & 255 \\ 0 & 255 & 255 & 255 & 255 \end{pmatrix} \quad \begin{pmatrix} 0 & 0 & 255 & 0 & 255 \\ 94 & 177 & 0 & 0 & 125 \\ 0 & 0 & 185 & 155 & 255 \\ 0 & 0 & 255 & 255 & 255 \\ 0 & 255 & 255 & 255 & 255 \end{pmatrix}$$

Corrupted image segment　　　　Restored image segment

**Case (b):** Initialize forward counter F=1 and reverse counter L=9. Convert the 2D array into 1D array and sort the converted array. When a 0 is detected F is incremented by 1 and when a 255 is detected L is decremented by 1.

　Unsorted array:　0　185　255　0　0　255　125　255　255
　Sorted array $S_{xy}$　　0　0　0　125　185　255 255　255

Here the median $S_{med}$ value is 185. The case (2) is illustrated as follows. Now check for the presence of 0 or 255 in the sorted array. Every time a 0 is detected F is incremented by 1 and if 255 is detected L is decremented by1. In the above example there is three 0 and three 255. Hence F is incremented by three times and L is decremented by three times. Now finally F is holding 4 and L is holding 6. Now the variable DET is assigned with the median of the rank ordered unsymmetrical trimmed output i.e. corrupted pixel is replaced by median (125,185) = 155. i.e., DET=155. Now perform first step detection │0-155│ > 40. This condition is true. The Second condition is checked │185-155│ > 20 and the second condition is true. Hence the processed pixel and the computed median is considered as noisy. Hence the corrupted pixel is replaced with Unsymmetrical trimmed median ie 155 output=155.

$$\begin{pmatrix} 0 & 0 & 255 & 0 & 255 \\ 104 & 119 & 255 & 255 & 255 \\ 0 & 103 & 255 & 255 & 123 \\ 0 & 122 & 255 & 124 & 255 \\ 0 & 255 & 255 & 255 & 255 \end{pmatrix} \quad \begin{pmatrix} 0 & 0 & 255 & 0 & 255 \\ 104 & 119 & 255 & 255 & 255 \\ 0 & 103 & 255 & 255 & 123 \\ 0 & 122 & 255 & 124 & 255 \\ 0 & 255 & 255 & 255 & 255 \end{pmatrix}$$

Corrupted image Segment      Restored image Segment

**Case (3):** Initialize F=1 and L=9. After sorting the current window in ascending order, the counters propagate in the 1D array resulting in holding count=6, F=4 and L=6. DET will hold median (103, 104, 119) ie, DET=104. Now perform impulse detection │119-104│ > 40. This condition is false and hence processed pixel is considered as non noisy hence left unaltered [13].

### 3.4 FPGA implementation of the proposed Algorithm

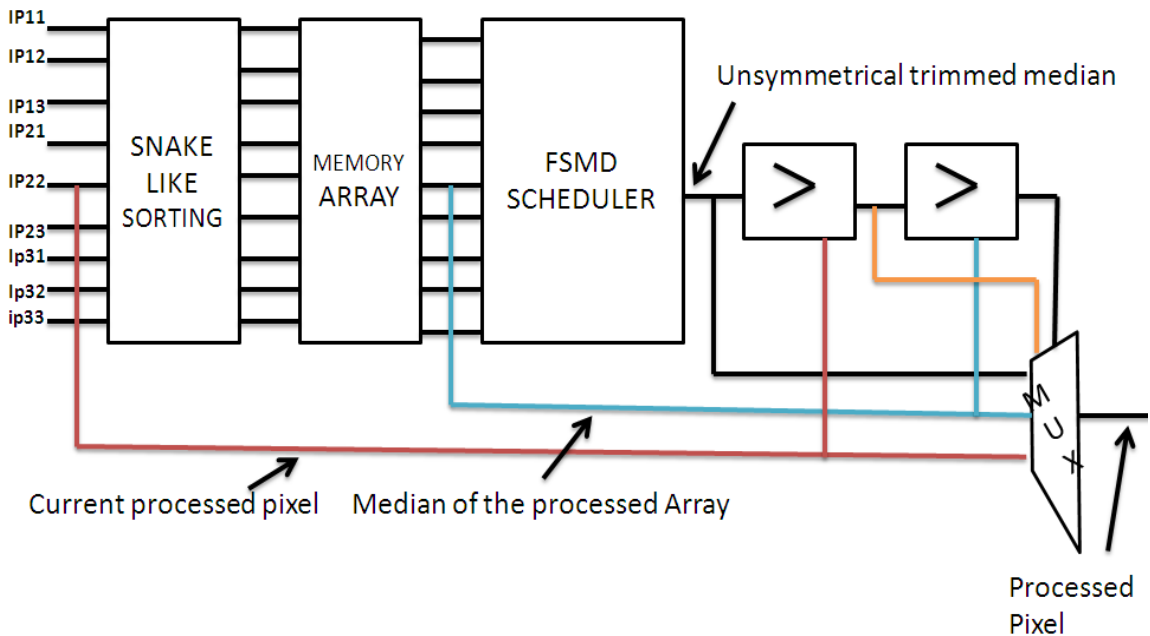

Figure 2 Proposed Sequential Architecture

The proposed algorithm is implemented for the FPGA device **Xc3e5000-5fg900** using VHDL The proposed sequential architecture consists of various snake like sorting, FSMD scheduler, Decision maker unit as shown in figure 2.

### 3.4.1 Snake like sorting:

We propose a parallel architecture for the proposed algorithm which uses 3x3 spatial windows for processing. The Proposed architecture is illustrated in figure 3. The Basic Processing element of the proposed architecture uses a three cell sorter. The function of the three cell sorter is order the data and produce outputs in maximum, middle and minimum values. The three pixel elements of the first, second and third rows are sent inside the three parallel three cell sorter as part of arranging first and third row in ascending and the second one is descending. This results in minimum1, minimum2, minimum3, middle1, middle2, middle3, maximum1, maximum2, maximum3. In the second phase minimum 1, maximum2 and minimum3 is fed to the first of three cell sorters, middle1, middle2, middle3 is given to second three cell sorters. The maximum1, minimum2 and maximum3 are fed to the third three cell sorter. This result in column sorting thereby the sorted signals are given as minimum4, minimum4, minimum4 middle5, middle5, middle5, maximum4, maximum5, maximum6. For the second row sorting minimum4,5,6 is fed into first three cell sorter, middle 4,5,6,maximum 4,5,6 into second and third level of three cell sorters respectively. The output signals of this stage are marked as minimum7, minimum8, minimum9 middle7, middle8, middle9, maximum7, maximum8, maximum9. The second column sorting is facilitated by minimum 7, maximum8 and minimum9 is fed to the first of three cell sorters, middle7, middle8, middle9 is given to second three cell sorters. The maximum7, minimum8 and maximum9 are fed to the third three cell sorter. The resultant signal is denoted as minimum10, minimum11, minimum12 middle10, middle11, middle12, maximum10, maximum11, maximum12. Now to facilitate the semi diagonal sorting minimum11,median11,maximum11 are given to the first of last stage three cell sorters and middle10,maximum10,maximum11is given to second of the last three cell sorter. The output is marked as min1, min2, min3 and max1, max2, max3 from the last stage sorters respectively. Minimum10 is the minimum value of the array, max1, 2, 3 refers to the second, third, fourth minimum of an array. Middle 11 is marked as the median of the array. max1, 2, 3 gives the maximum values after median in the array. Maximum 12 is maximum value of the array. The order specifies the rank ordering of the array [14].

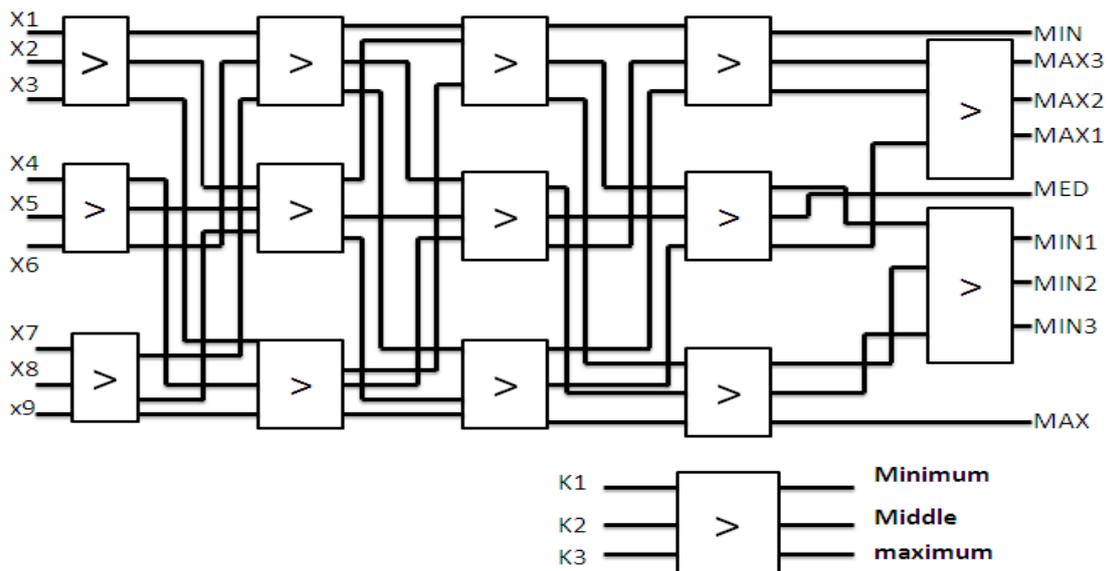

Figure 3: Parallel architecture for the proposed Snake like sorting algorithm

### 3.4.2 FSMD Scheduler Unit
The proposed FSMD scheduler unit consists of 8 states. The states are named as idle, Dat1, index, Decision, Out_even, Out_odd, final process, output final. When the system reset is inactive the control of the program is transferred to the next state called idle.

### 3.4.2.1 Idle State:
The counters such as F, L, (noise determination counter) and the sum accumulator is initialized to zero and points to Dat1 as next state.

### 3.4.2.2 DAT1 State:
Every element of the 3X3 is checked for 0 or 255. If 0 is encountered counter F is incremented by 1. If 255 are encountered L is incremented by 1. The total number of noisy pixel in the array is stored to a variable called t_noise. When the counters F and L reaches the maximum value as 9 the program control is transferred to next state called index.

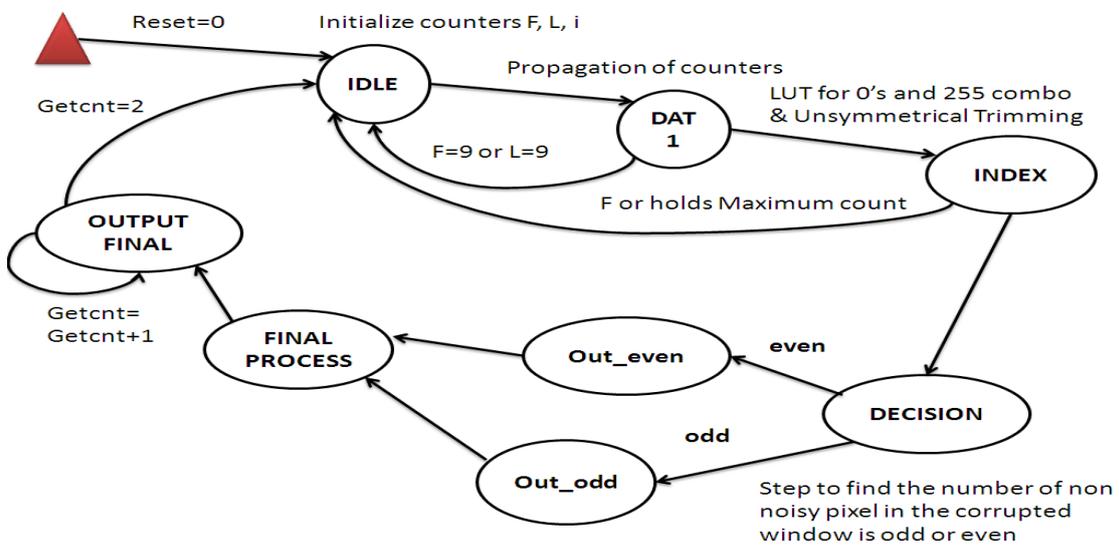

Figure 4 FSMD Scheduler for evaluation of Unsymmetrical trimmed median filter

### 3.4.2.3 Index State:
In this state two conditions are checked. First, when all the elements of the 3x3 window are combination of 0 or 255 then a Look up table is formulated such that if all the elements are 0 then variable op is 0. If all the values are 255 then op is 255. If both the above condition is failed then Look up table is loaded with 198,170,141,113,85,56,28 indicating the mean of the elements in an array with 2 0's, 3 0's 4 0's 5 0's 6 0's 7 0's 8 0's respectively (i.e., sum of all 255 divided by 9). On the second case, when all the elements are not the combination of 0 or 255 then non zero entries index is checked. This state gives the procedure to point index of the elements to find unsymmetrical trimmed median, if it is odd or even depending upon the number of noisy pixel within the given window. The logic is to prefix certain index in the form of look up table so that for an example when no 0's and three 255's is present then the number of non noisy pixel is even i.e. 6. Index for finding the median is fixed as 3 and 4. In the case of four 255's, then the number of non noisy pixel is odd i.e. 5. So the index is prefixed as 3. Similarly the index positions are prefixed in look up table for all possible combination. Depending upon the number of noisy pixels in the given window the values are accessed from the look up table. If the number of noisy elements are even then the index are stored in even_u and even_v respectively. If the number of noisy elements are odd then the index are stored in odd. After finding the index the next state is pointed to Decision state.

**3.4.2.4 Decision State**
After obtaining the index from the index state the value corresponding to the index state is obtained. Initially the number of noisy pixels is evaluated and based on the noisy pixel the unsymmetrical trimmed median is obtained. If the number of non-noisy pixel is odd then sum is obtained as (memory (even_u) +memory (even_v)). In case of odd number, the non noisy pixel is even then unsymmetrical trimmed median is obtained as memory (odd). If the number of non noisy pixel is odd then the next state is pointed as out_even. If the number of non noisy pixel is even then the next state pointed as out_odd.

**3.4.2.5 Out_even State**
After finding the number of noisy pixel as even the control is transferred to out_even state. In this state the unsymmetrical trimmed median is obtained by finding the mean of memory (even_u) and memory (even_v) and points to the next state called final process.

**3.4.2.6 Out_odd State**
After finding the number of noisy pixel as odd the control is transferred to out_odd state. In this state the unsymmetrical trimmed median is obtained by finding the odd$^{th}$ element of the trimmed array i.e. memory (odd) and points to the next state called final process.

**3.4.2.7 Final Process State:**
The centre pixel, Unsymmetrical trimmed median value and median of the sorted array is loaded into this state for the final process. This stage is done to obtain synchronization between the output variables. The next state is pointed as out_final.

**3.4.2.8 Out_final State:**
Here the decision to check the centre pixel is noisy as follows. If the absolute difference between centre pixel and unsymmetrical trimmed median is greater than 40. If the condition is true then the processed pixel is considered as noisy. Now check for the median is noisy or not by finding the absolute difference between computed median and unsymmetrical trimmed median is greater than 20 then the computed median is noisy. Hence the processed pixel is replaced with unsymmetrical trimmed median else if the median is not noisy then replace the processed pixel with median of the array. If the centre pixel is not noisy it is left unaltered.

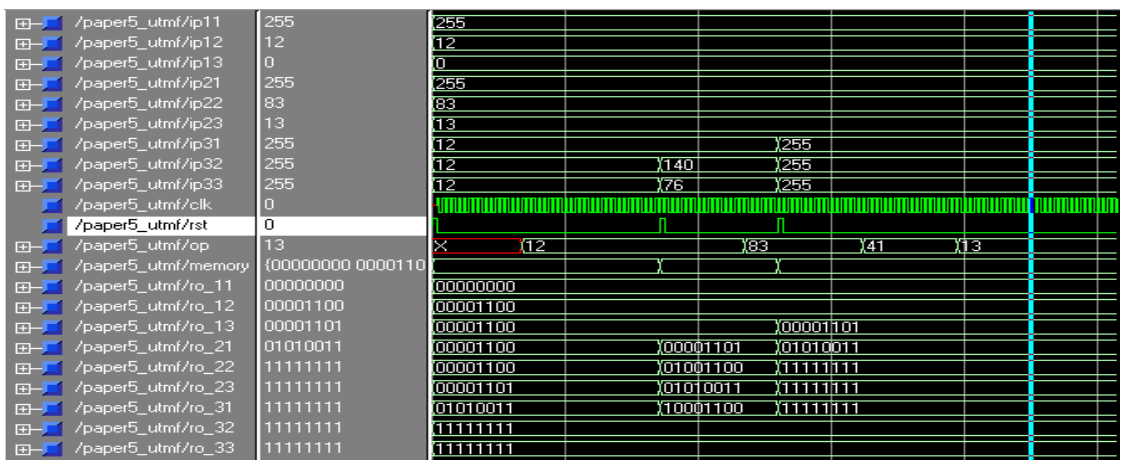

**Figure 5. Simulation results of the architecture of Proposed algorithm**

TABLE I

PERFORMANCE OF VARIOUS ALGORITHMS AT DIFFERENT FIXED VALUED IMPULSE NOISE DENSITIES FOR PSNR AND IEF IN LENA IMAGE

| ND | PSNR | | | | | | | IEF | | | | | | |
|---|---|---|---|---|---|---|---|---|---|---|---|---|---|---|
| | SMF | AMF | CWF | TDF | PSM | DPF | PA | SMF | AMF | CWF | TDF | PSM | DPF | PA |
| 10% | 34.9 | 39.3 | 35.2 | 32.7 | 38.8 | 33.8 | 39.0 | 89 | 246 | 95.9 | 38.2 | 219 | 69.1 | 229.3 |
| 20% | 30.3 | 36.9 | 28.1 | 27.8 | 33.4 | 27.5 | 35.7 | 61 | 281 | 37.2 | 25 | 124 | 32.4 | 217.2 |
| 30% | 23.9 | 34.6 | 22.2 | 23.3 | 29.4 | 23.1 | 33.6 | 21.4 | 254 | 14.4 | 19.6 | 74.5 | 17.8 | 200.3 |
| 40% | 19.0 | 32.2 | 17.8 | 19.0 | 25.4 | 19.7 | 31.7 | 9.1 | 192 | 6.9 | 9.2 | 40.1 | 10.7 | 169.4 |
| 50% | 15.9 | 27.3 | 14.3 | 15.3 | 25.3 | 16.8 | 30.1 | 4.9 | 78 | 3.9 | 4.8 | 39.6 | 6.8 | 148.2 |
| 60% | 12.3 | 21.6 | 11.7 | 12.4 | 21.2 | 14.5 | 28.1 | 2.9 | 25 | 2.5 | 2.9 | 19.1 | 4.8 | 112.9 |
| 70% | 10.0 | 16.6 | 9.6 | 10.0 | 9.9 | 12.5 | 25.8 | 2.0 | 9.1 | 1.8 | 2 | 1.9 | 3.5 | 77.6 |
| 80% | 8.1 | 12.7 | 7.9 | 8.1 | 8.1 | 10.7 | 23.2 | 1.4 | 4.3 | 1.4 | 1.4 | 1.4 | 2.7 | 47.9 |
| 90% | 6.6 | 9.8 | 6.5 | 6.6 | 6.6 | 9.2 | 19.2 | 1.1 | 2.5 | 1.1 | 1.1 | 1.1 | 2.1 | 21.8 |

TABLE II

PERFORMANCE OF VARIOUS ALGORITHMS AT DIFFERENT RANDOM VALUED IMPULSE NOISE DENSITIES FOR PSNR AND IEF IN BABOON IMAGE

| ND | PSNR | | | | | | IEF | | | | | |
|---|---|---|---|---|---|---|---|---|---|---|---|---|
| | SMF | AMF | MEAN DET | MED DET | CUM TPF | PA | SMF | AMF | MEAN DET | MED DET | CUM TPF | PA |
| 10% | 23.92 | 23.26 | 28.48 | 27.89 | 23.30 | 28.65 | 2.6 | 2.1 | 7.6 | 6.56 | 1.2 | 7.5 |
| 20% | 23.59 | 19.27 | 26.13 | 26.15 | 21.32 | 26.61 | 4.8 | 1.7 | 8.4 | 8.47 | 1.4 | 9.6 |
| 30% | 23.08 | 17.07 | 23.98 | 24.60 | 19.60 | 25.10 | 6.4 | 1.5 | 8.0 | 8.98 | 1.6 | 10.0 |
| 40% | 22.72 | 15.35 | 22.26 | 22.78 | 18.40 | 23.45 | 7.6 | 1.4 | 7.0 | 7.88 | 2.8 | 9.1 |
| 50% | 21.84 | 16.32 | 20.55 | 20.73 | 17.41 | 22.16 | 7.9 | 2.2 | 5.8 | 6.19 | 2.8 | 8.5 |
| 60% | 20.86 | 15.00 | 18.85 | 19.02 | 16.73 | 20.85 | 7.6 | 1.9 | 4.7 | 4.99 | 2.9 | 7.5 |
| 70% | 19.57 | 14.00 | 17.50 | 17.33 | 16.16 | 19.15 | 6.5 | 1.8 | 4.0 | 3.94 | 3.0 | 5.9 |
| 80% | 17.87 | 13.06 | 15.99 | 15.77 | 15.67 | 17.17 | 5.0 | 1.6 | 3.3 | 3.15 | 3.1 | 4.9 |
| 90% | 16.50 | 12.23 | 14.67 | 14.64 | 15.31 | 15.73 | 4.1 | 1.5 | 2.76 | 2.73 | 3.2 | 3.5 |

TABLE III

PERFORMANCE OF VARIOUS ALGORITHMS AT DIFFERENT FIXED AND RANDOM VALUED IMPULSE NOISE DENSITIES FOR MSE IN LENA AND BABOON IMAGE.

| ND In % | MSE(FVIN) | | | | | | | MSE(RVIN) | | | | | |
|---|---|---|---|---|---|---|---|---|---|---|---|---|---|
| | SMF | AMF | CWF | TDF | PSM | DPF | PA | SMF | AMF | MEAN DET | MED DET | CUM TPF | PA |
| 10 | 20 | 7 | 20 | 26 | 8 | 27 | 8 | 263 | 306 | 92 | 105 | 303 | 88 |
| 20 | 60 | 13 | 102 | 105 | 29 | 114 | 17 | 284 | 767 | 158 | 157 | 479 | 141 |
| 30 | 259 | 22 | 409.9 | 286 | 74 | 312 | 27 | 319 | 1276 | 259 | 225 | 711 | 200 |
| 40 | 814 | 38 | 1082 | 800 | 185 | 686 | 43 | 346 | 1892 | 385 | 342 | 937 | 293 |
| 50 | 1877 | 118 | 2367 | 1909 | 187 | 1355 | 62 | 424 | 1517 | 572 | 549 | 1178 | 395 |
| 60 | 3776 | 443 | 4295 | 3732 | 484 | 1197 | 98 | 532 | 2051 | 847 | 814 | 1378 | 533 |
| 70 | 6379 | 1421 | 7109 | 6450 | 600 | 3651 | 168 | 716 | 2588 | 1154 | 1200 | 1570 | 789 |
| 80 | 9945 | 3413 | 10624 | 9843 | 1000 | 5408 | 310 | 1061 | 3207 | 1633 | 1720 | 1760 | 1101 |
| 90 | 14179 | 6708 | 14513 | 13922 | 1396 | 7798 | 766 | 1453 | 3890 | 2214 | 2231 | 1913 | 1735 |

TABLE IV

PERFORMANCES OF VARIOUS ALGORITHMS AT DIFFERENT ZERO MEAN GAUSSIAN NOISE DENSITIES FOR PSNR AND MSE IN BABOON IMAGE.

| VAR | PSNR | | | | | | MSE | | | | | |
|---|---|---|---|---|---|---|---|---|---|---|---|---|
| | SMF | AMF | MEAN DET | MED DET | CUM TPF | PA | SMF | AMF | MEAN DET | MED DET | CUM TPF | PA |
| 0.001 | 24.0 | 27.5 | 27.8 | 26.8 | 24.7 | 27.73 | 257 | 115 | 105 | 134 | 219 | 109 |
| 0.002 | 23.8 | 26.3 | 25.9 | 25.4 | 24.5 | 25.72 | 266 | 151 | 167 | 187 | 228 | 173 |
| 0.003 | 23.7 | 25.4 | 24.6 | 24.5 | 24.3 | 24.55 | 273 | 185 | 223 | 228 | 236 | 228 |
| 0.004 | 23.6 | 24.7 | 23.8 | 23.8 | 24.2 | 23.67 | 277 | 219 | 267 | 267 | 245 | 279 |
| 0.005 | 23.5 | 24.1 | 23.31 | 23.4 | 24.1 | 23.04 | 286 | 252 | 303 | 295 | 252 | 322 |
| 0.006 | 23.4 | 23.5 | 22.9 | 23.0 | 23.9 | 22.59 | 291 | 289 | 331 | 320 | 261 | 358 |
| 0.007 | 23.3 | 23.1 | 22.5 | 22.7 | 23.8 | 22.21 | 300 | 317 | 362 | 343 | 269 | 390 |
| 0.008 | 23.2 | 22.6 | 22.3 | 22.6 | 23.6 | 21.88 | 304 | 353 | 380 | 356 | 279 | 420 |
| 0.009 | 23.2 | 22.3 | 22.0 | 22.3 | 23.5 | 21.64 | 307 | 379 | 404 | 381 | 288 | 445 |

TABLE V

QUANTITATIVE PERFORMANCE OF VARIOUS EXISTING FILTERS CORRUPTED BY 70 % FIXED VALUE IMPULSE NOISE FOR LENA IMAGE

| S.No | NAME OF THE FILTER | PSNR | IEF | MSE |
|---|---|---|---|---|
| 1 | SMF(3X3) | 10.07 | 2.03 | 6396 |
| 2 | SMF(5X5) | 14.23 | 5.32 | 2449 |
| 3 | AMF | 16.73 | 9.40 | 1379 |
| 4 | CWMF | 9.63 | 1.83 | 7077 |
| 5 | TDF | 10.03 | 2.02 | 6452 |
| 6 | MEAN DET | 10.11 | 2.05 | 6325 |
| 7 | MED DET | 10.08 | 2.03 | 6373 |
| 8 | RWCWMF | 9.61 | 1.83 | 7109 |
| 9 | PSMF | 9.98 | 1.99 | 6525 |
| 10 | DPF | 12.45 | 3.53 | 3691 |
| 11 | DBA | 27.80 | 120.93 | 107.80 |
| 12 | CDMUTMF | 21.81 | 2.48 | 38.75 |
| 13 | CUTMF | 30.79 | 240.94 | 54.13 |
| 14 | CUTMPF | 30.20 | 210.33 | 61.98 |
| 15 | MDBUTMF | 29.68 | 186.37 | 69.90 |
| 16 | PROPOSED ALGORITHM | 25.97 | 78.76 | 164.18 |

TABLE VI

TABLE PERFORMANCE OF PROPOSED ALGORITHM ON VARIOUS IMAGES FOR MIXED NOISE (30% FIXED VALUE IMPULSE NOISE PLUS ZERO VARIANCE 0.001 VARIANCE GAUSSIAN NOISE)

| S.no | IMAGE | PSNR | IEF | MSE |
|---|---|---|---|---|
| 1 | Lena | 28.61 | 60.44 | 89.36 |
| 2 | Barbara | 24.16 | 21.75 | 249.19 |
| 3 | Baby | 27.55 | 55.21 | 114.26 |
| 4 | Cameraman | 23.94 | 22.39 | 261.96 |
| 5 | Baboon | 25.35 | 26.65 | 189.44 |
| 6 | Peppers | 27.20 | 45.24 | 123.66 |

TABLE VII
COMPUTATION TIME OF DIFFERENT SORTING TECHNIQUES IMPLEMENTED IN MATLAB7 (R14) PENTIUM DUAL CPU E2140 @1.6 GHZ OF 1 GB RAM

| S.No | SORTING | TIME in sec |
|---|---|---|
| 1 | Bubble sorting | 17.07 |
| 2 | Insertion sorting | 20.46 |
| 3 | Quick Sorting | 139.76 |
| 4 | Selection Sorting | 18.26 |
| 5 | MDF Sorting | 58.203 |
| 6 | MATLAB in built function | 13.06 |
| 7 | Proposed Algorithm | 26.71 |

TABLE VIII
PERFORMANCE OF PROPOSED SNAKE LIKE SORTING ALGORITHM OVER CONVENTIONAL ALGORITHMS TARGETED ON **Xc3e5000-5fg900**

| S.no | PARAMETERS | BUBLE SORT | HEAP SORT | INSERTION SORT | MDF | SELECTION SORT | TDF | PA |
|---|---|---|---|---|---|---|---|---|
| | | | | AFTER SYNTHESIS | | | | |
| 1 | Slices | 4375 | 3810 | 4375 | 4021 | 4375 | 4132 | 709 |
| 2 | 4 I/P LUT | 6080 | 5312 | 6080 | 6854 | 6080 | 7066 | 1033 |
| 3 | Bonded IOB | 328 | 321 | 321 | 82 | 321 | 82 | 144 |
| 4 | Combinational Delay paths (ns) | 151.715 | 327.5 | 151.715 | 188.933 | 151.715 | 190.94 | 77.307 |
| | | | | AFTER MAPPING | | | | |
| 5 | Gate count | 43075 | 37,699 | 43,075 | 42,055 | 43,075 | 43,927 | 7,281 |
| | | | | AFTER PLACE AND ROUTE | | | | |
| 6 | Slices flip flop | 3088 | 2783 | 3088 | 3689 | 3088 | 3800 | 517 |
| | | | | POWER CONSUMED | | | | |
| 7 | Power consumption | 298 | 100 | 100 | 298 | 100 | 298 | 298 |

TABLE IX
PERFORMANCE OF 3X3 WINDOW OF PROPOSED ALGORITHM TARGETED ON **Xc3e5000-5fg900**

| S.No | PARAMETERS | PA |
|---|---|---|
| | AFTER SYNTHESIS | |
| 1 | Slices | 1034 |
| 2 | 4 I/P LUT | 116 |
| 3 | Bonded IOB | 1599 |
| 4 | Operating Frequency (MHz) | 79.935 |
| | AFTER MAPPING | |
| 5 | Gate count | 11945 |
| | AFTER PLACE AND ROUTE | |
| 6 | Slices flip flop | 116 |
| | POWER CONSUMED | |
| 7 | Power consumption mw | 298 |

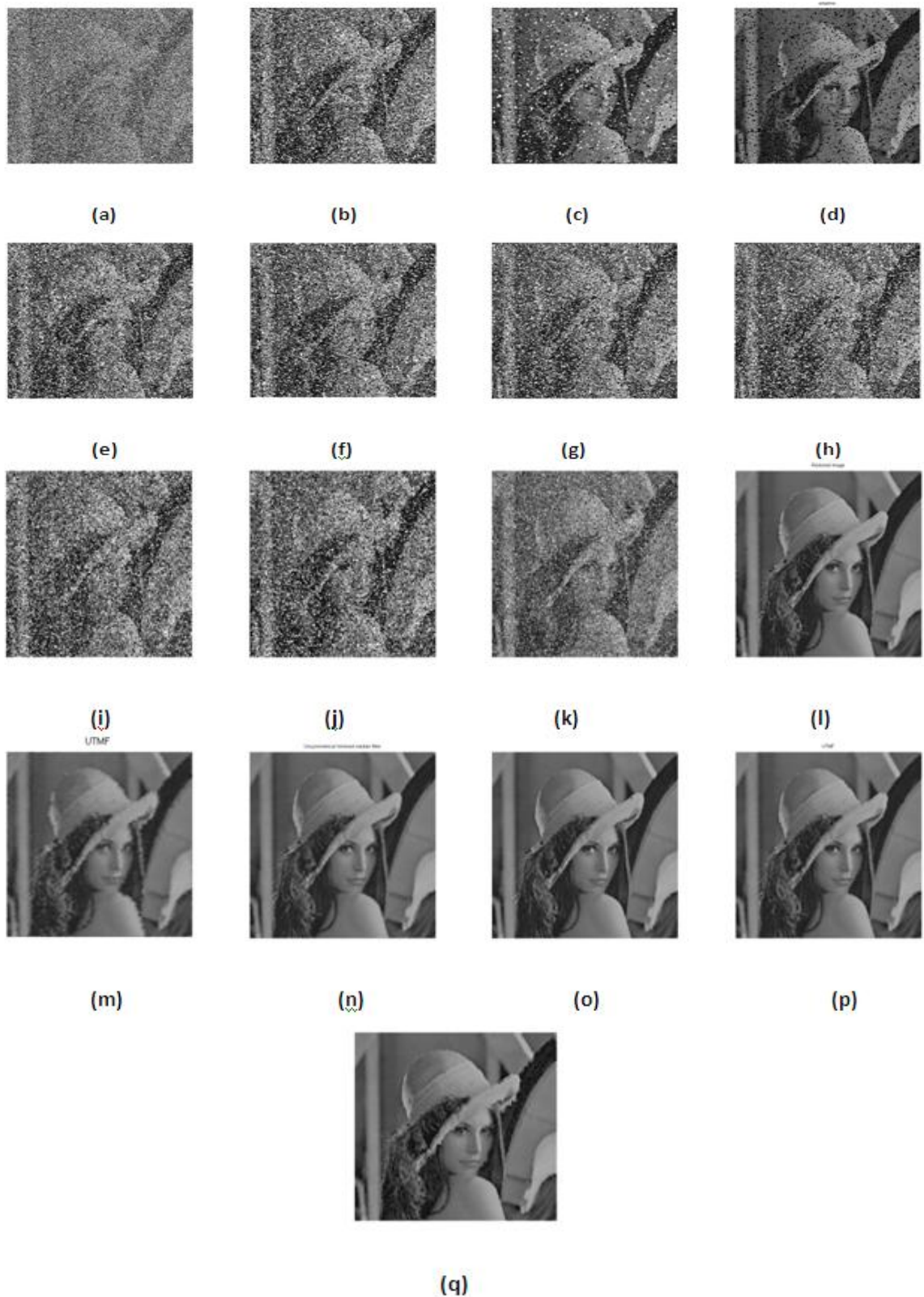

Figure 6 Qualitative performance of Various algorithm For 70% Fixed value impulse noise a) Corrupted image b) Smf(3x3)  c)Smf(5x5) d)AMF e) CWF f)TDF g)Mean Det h) Med Det i) RWCWMF j) PSMF k) DPF l) DBA  m) CDMUTMF n) CUTMF o) CUMTPF p) MDBUTMF q) PA

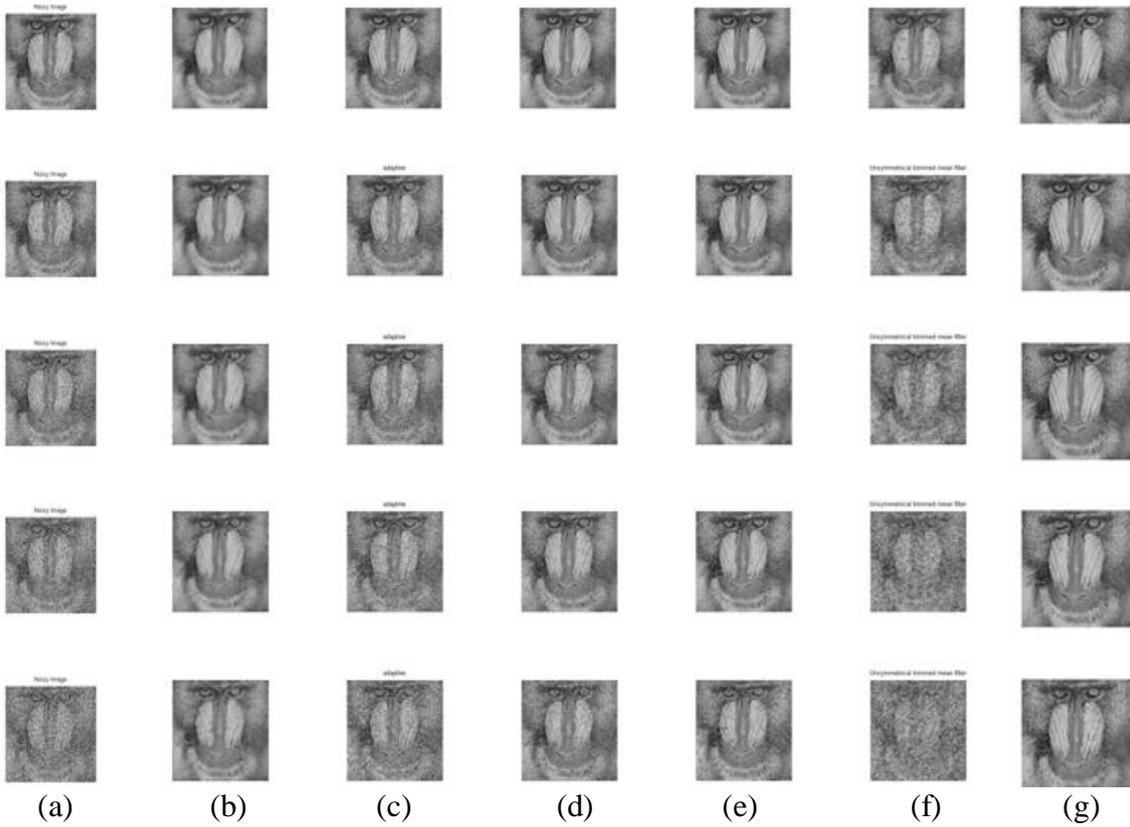

(a) (b) (c) (d) (e) (f) (g)

Figure 7. Performance of various filters for Baboon image corrupted by Random Valued Impulse noise from 10% to 50% in row1 to 5 respectively. Output of various filters in column 1 to 8 (a) Random valued impulse noise (b) output of SMF (c) output of AMF (d) output of Meandet (e) output of Meddet (f) output of CUMTF (g) output of PA

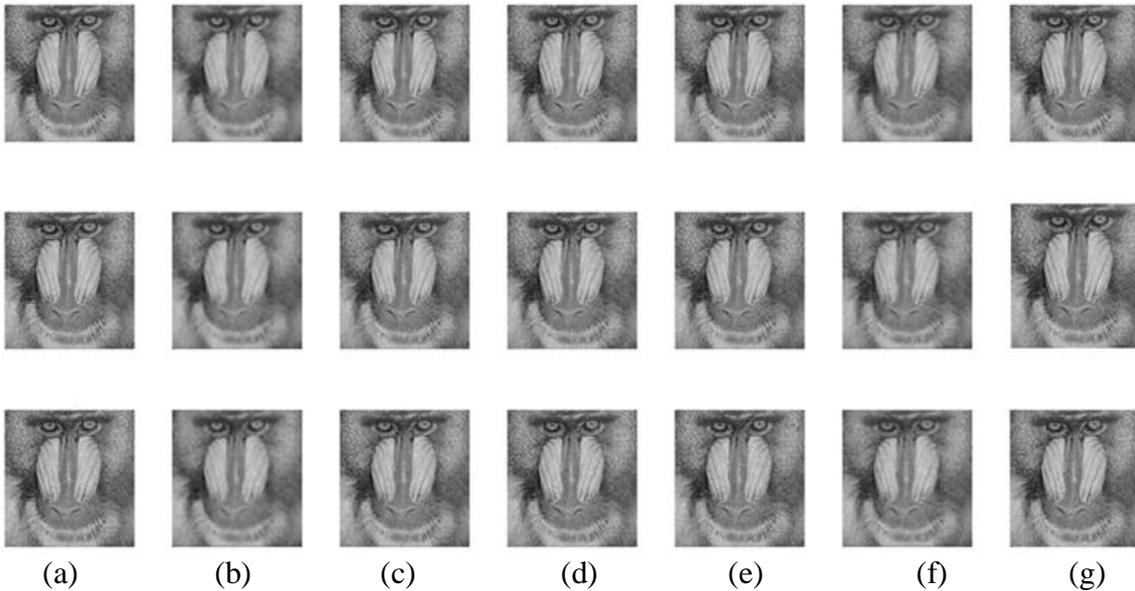

(a) (b) (c) (d) (e) (f) (g)

Figure 8. Performances of various filters for Baboon image corrupted by Zero mean Gaussian noise variance from 0.001 to 0.003 in row1 to 3 respectively. Output of various filters in column 1 to 7 (a) Random valued impulse noise (b) output of SMF (c) output of AMF (d) output of Meandet (e) output of Meddet (f) output of CUMTF (g) output of PA

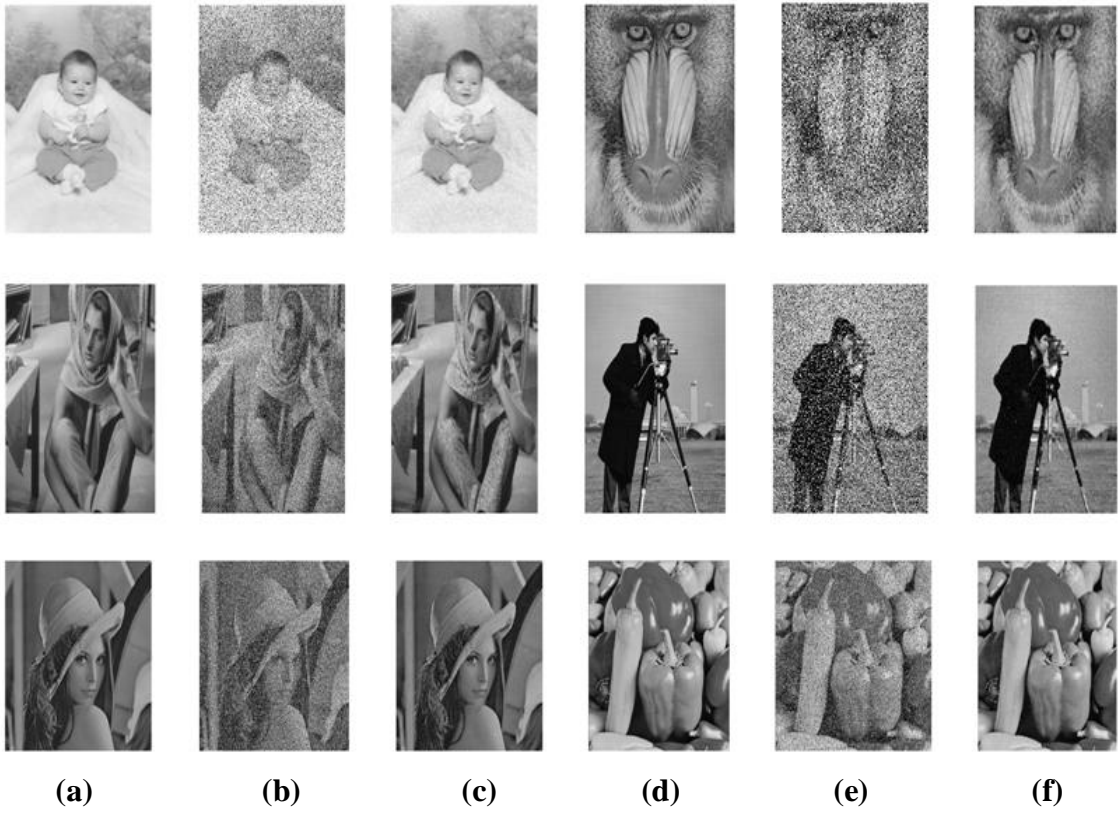

|  (a)  |  (b)  |  (c)  |  (d)  |  (e)  |  (f)  |

Figure 9. Performances of Proposed algorithm for various image corrupted by mixed noise (30% impulse noise plus zero mean Gaussian noise variance of 0.001 in row1 to 3 respectively. (a & d) original image (b & e) Mixed noise (c & f) output of PA.

## 4. SIMULATION RESULTS & DISCUSSIONS

The Quantitative performance of the proposed algorithm is evaluated based on Peak signal to noise ratio (PSNR), Mean Square Error (MSE) and Image Enhancement Factor (IEF) which is given in equations 4,5,6 respectively.

$$\text{PSNR} = 10\log_{10}\left(\frac{255^2}{MSE}\right) \quad (4)$$

$$\text{MSE} = \frac{\sum_i \sum_j (r_{ij} - x_{ij})2}{M \times N} \quad (5)$$

$$\text{IEF} = \frac{\left(\sum_i \sum_j n_{ij} - r_{ij}\right)^2}{\left(\sum_i \sum_j x_{ij} - r_{ij}\right)^2} \quad (6)$$

Where *r* refers to Original image, n gives the corrupted image, *x* denotes restored image, M x N is the size of Processed image [13]. The existing algorithms used for the comparison are SMF, AMF, CWF, TDF, PSMF, DPF for comparing Fixed value impulse noise. To compare random valued impulse noise and Gaussian noise, the algorithms such as SMF, AMF, MEANDET, MEDDET is used. The qualitative performance of the proposed algorithm is tested on various images such as Lena,

Cameraman, Baboon, Barbara, girl, pepper etc (Images are chosen as per the details of the image). Quantitative analysis is made by varying noise densities in steps of ten from 10% to 90% for Random valued impulse noise (RVIN) and Fixed valued impulse noise (FVIN). The same is done by varying variance of zero mean Gaussian noise in increment of 0.001. The proposed algorithm is also tried for 70% of Fixed valued impulse noise and mixed noise on low detail, medium detail and high detail images. Comparisons were made in terms of PSNR, IEF and MSE. Results and graphs are given in Table I-VI and figure 10-17 respectively. Figure 6-9 gives the qualitative performance of the proposed algorithm in terms of noise elimination for FVIN, RVIN, Zero mean Gaussian noise, and mixed noise. All the simulation is done in dual CPU E2140@1.6Ghz with 1GB RAM capacity. Better results were obtained when the pre-defined threshold T was between 20 and 40. And the second threshold T1 was between 15 and 30. From the Table I we infer that for the proposed algorithm has high PSNR and IEF, indicating how much the algorithm eliminates salt and pepper noise effectively. Table II gives the performance of random valued impulse noise of various algorithms. The proposed algorithm is found to work well for low density RVIN noise. From Table III we find the mean square error is minimum for proposed algorithm at high noise densities for both FVIN and RVIN. It is evident from figure 6 that the qualitative aspect of the proposed algorithm at 70% FVIN is found to perform good against conventional algorithms. The proposed algorithm performs on par with the recently proposed decision based filters for Lena (low detail image). Table IV and V gives the quantitative performance of zero mean Gaussian noise and 70% FVIN respectively. It is shown that the proposed algorithm fairs good for all these noises. Figure 7 gives the performance of various filters for the baboon image corrupted by RVIN from 10% to 50%. It was found that the existing algorithm either blurs the image or fails to remove the noise. The proposed algorithm is effective in removing RVIN for noises up to 30%. Many traditional algorithms do not perform well for high noise densities. Hence none of the single level detector algorithm is able to detect and correct the long tailed noise at high noise densities. From table IV and figure 8 we understand that the proposed algorithm fairs well for zero mean Gaussian noise for increasing variance. Table VI and Figure 9 illustrates the performance of proposed algorithm for mixed noise (30% impulse noise and zero mean 0.001 variance of Gaussian noise). It is found to perform well in eliminating mixed noise also. The value of the threshold is updated based on the number of corrupted pixels inside the corrupted window. Table 7 gives the computation time of different sorting technique implemented in Matlab 7 on Pentium dual cpu E2140 @1.6 GHz of 1 GB RAM and found to perform on par with various sorting algorithms. Figure 10 to 17 illustrates the graphical performance of the proposed algorithm. The results illustrates that the proposed algorithm has a good PSNR,IEF for high density FVIN noise, optimum results for high density Random valued impulse noise, good quantitative measure on zero mean Gaussian noise. The proposed algorithm was targeted on Spartan 3e family XC3S5000-5fg900 FPGA. The code was developed using VHDL. The simulator tools used was a third party tool Modelsim 5.8i and synthesis tool XST was used as part of Xilinx 7.1i suit for CPLD & FPGA development. Table 8 gives the device utilization summary, timing specification and power report for the target FPGA for various median finding algorithms such as bubble sort, heap sort, insertion sort, Selection sort, Threshold decomposition Filter. Table 9 illustrates the performance of proposed algorithm for the targeted device. Figure

5 gives the simulation output of the PA which is found to generate its first output in 13 clock cycles.

## 5. CONCLUSION

All the algorithms were tested on a fixed 3x3 window. From the exhaustive experiments, we conclude that the proposed algorithm has a high PSNR, low MSE and high IEF for different images and for different noise type at higher noise densities. However, on an average sense, PA gives good performance in eliminating FVIN up to 70%, RVIN up to 30%, zero mean 0.5% variance Gaussian noise and a mixed noise (30% impulse noise plus zero mean 0.001 variance Gaussian noise). When compared to Conventional filters such as SMF, AMF, CWF, TDF,PSMF,DPF etc , the PA exhibits good performance for Salt & Pepper noise removal up to 70% and reduces smaller proportion of zero mean 0.3% variance Gaussian noise. The proposed filter also exhibits good noise removal up to 30% RVIN and 30% of mixed noise. The proposed algorithm works on par with the recently proposed algorithms such as DBA, MDBUTMF, CUTMF etc., in our method, time complexity of the existing methods is eliminated by using the pixel intensity itself as threshold. Hence, the proposed method shows optimum performance with fewer comparison complexities. The Proposed algorithm has good average computation time. FPGA implementation of the proposed algorithm for 3x3 window is implemented and performance in terms of area, speed and power is illustrated in table 8, 9 respectively. Table 8 gives the device utilization summary, timing and power specification for the target device XC3S5000-5fg900 required by the snake like algorithm with the existing sorting algorithm. The proposed snake like algorithm utilizes 709 slices, which is 60% less when compared to other sorting algorithms. The snake like sorting also has a low combinational delay path of 77.30ns with a reduced gate count and slices flip flop of 7281 and 517 respectively, which is 7 times less when compared to existing algorithm. The last part of the table deals with power required by each sorting algorithm on the FPGA. The proposed logic for the entire algorithm is implemented on the FPGA and found to consume 1034 slices with an operating frequency of 79.93 MHz and a gate count of 11945 with optimum power consumption of 298mw. It was found that the proposed parallel snake like sorting logic requires very less area and time with optimum power consumption when compared to the existing sorting techniques. The proposed algorithm exhibits very good results in restoration of images corrupted by non identical noise both quantitatively and qualitatively and occupies a low area, good operating frequency and optimum power architecture is proposed.

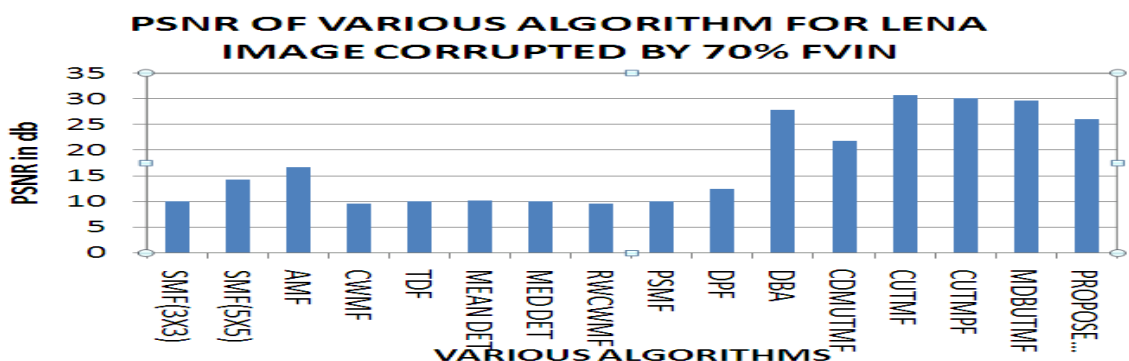

Figure 10. PSNR of various algorithms for Lena image corrupted by 70% FVIN

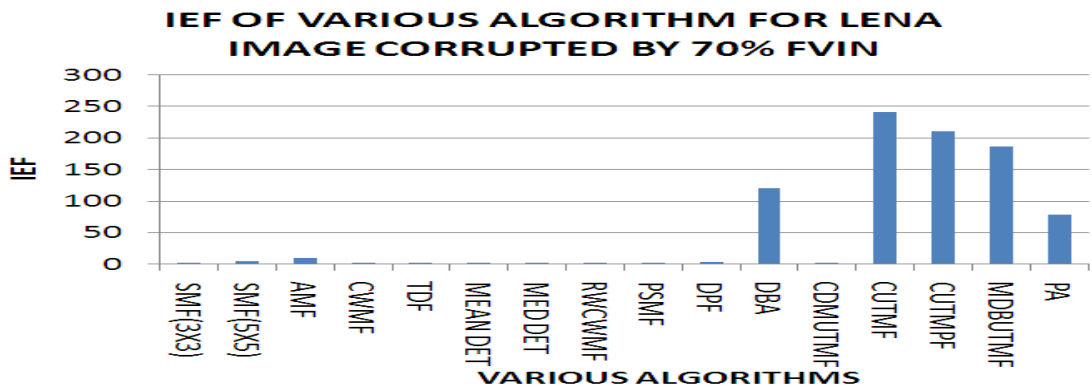

Figure 11. IEF of various algorithms for Lena image corrupted by 70% FVIN

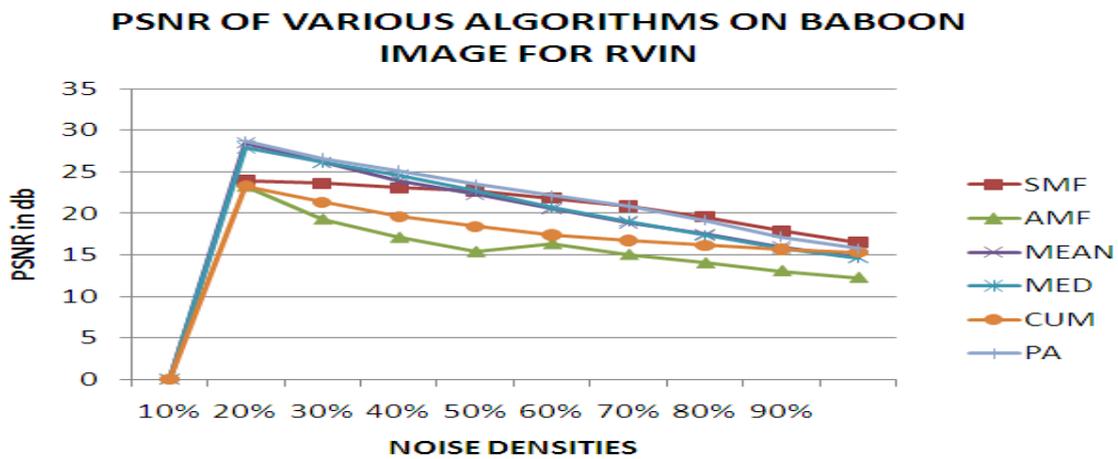

Figure 12. PSNR of various algorithms for BABOON image corrupted by RVIN

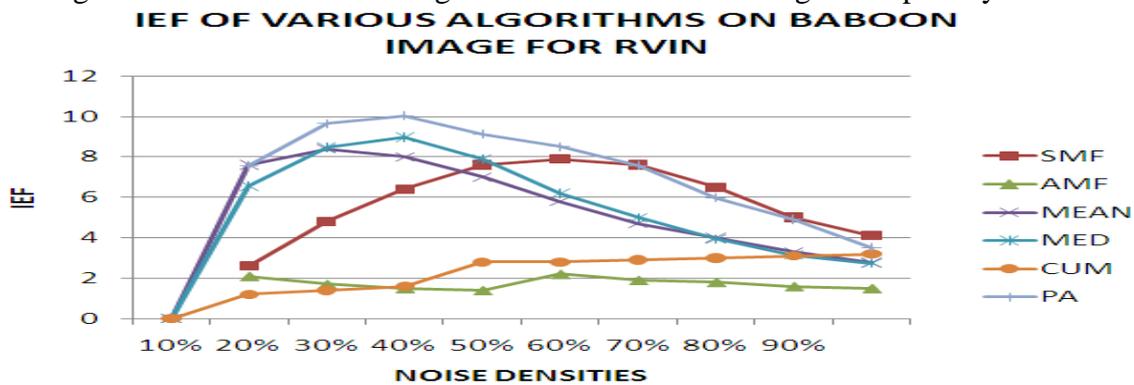

Figure 13. IEF of various algorithms for BABOON image corrupted by RVIN

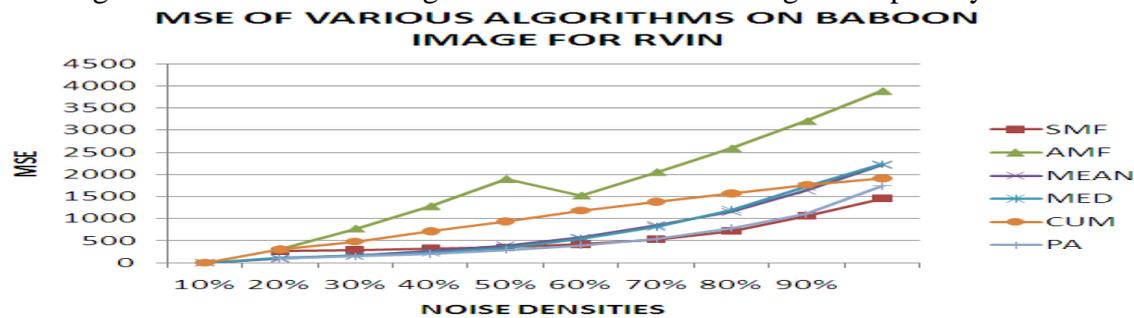

Figure 14. MSE of various algorithms for BABOON image corrupted by RVIN

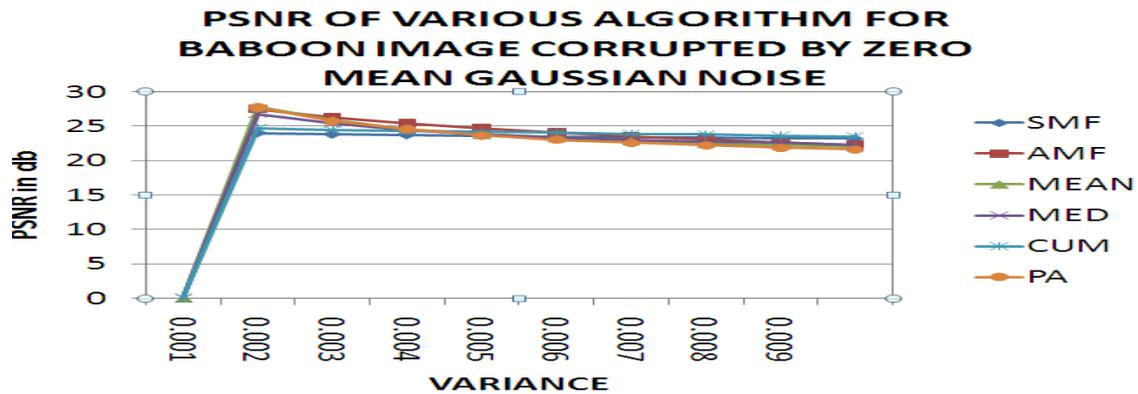

Figure 15. PSNR of various algorithms for BABOON image corrupted by Zero mean Gaussian noise

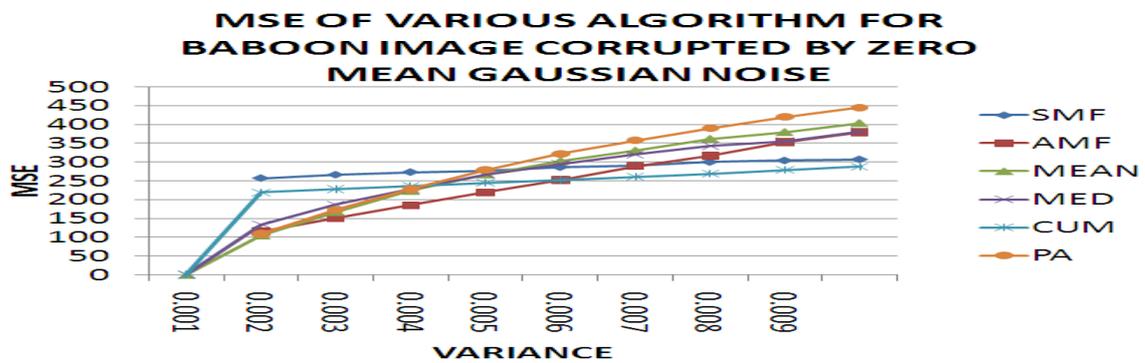

Figure 16. MSE of various algorithms for BABOON image corrupted by Zero mean Gaussian noise

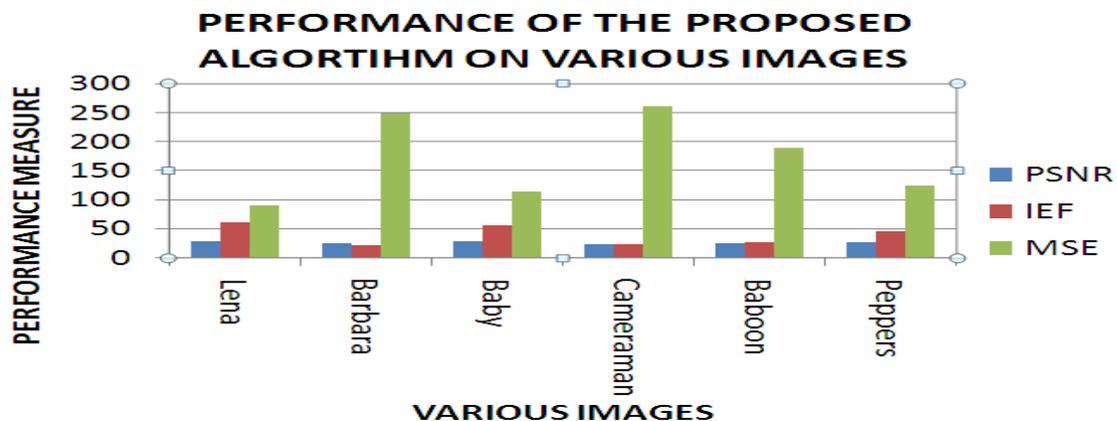

Figure 17. PSNR, IEF, MSE of various algorithms for images corrupted by Mixed noise